\pdfoutput=1
\documentclass[10pt,twocolumn,letterpaper]{article}

\usepackage{cvpr}
\usepackage{times}
\usepackage{epsfig}
\usepackage{graphicx}
\usepackage{amsmath}
\usepackage{amssymb}
\usepackage[hidelinks]{hyperref}

\graphicspath{ {images/} }
\usepackage{caption}
\usepackage{subcaption}
\usepackage[utf8]{inputenc}
\usepackage{color}
\usepackage{array}
\usepackage{epstopdf}
\newtheorem{theorem}{Theorem}
\usepackage{tikz}
\usetikzlibrary{tikzmark}

\cvprfinalcopy 

\def\cvprPaperID{****} 

\begin{document}

\title{ Revisiting Winner Take All (WTA) Hashing for Sparse Datasets}

\author{Beidi Chen\\
    Department of Computer Science \\
	Rice University\\
	{\tt\small beidi.chen@rice.edu}
	\and
	Anshumali Shrivastava\\
    Department of Computer Science\\
	Rice University\\
	{\tt\small anshumali@rice.edu}
}

\maketitle

\begin{abstract}
WTA (Winner Take All) hashing has been successfully applied in many large scale vision applications. This hashing scheme was tailored to take advantage of the comparative reasoning (or order based information), which showed significant accuracy improvements.
In this paper, we identify a subtle issue with WTA, which grows with the sparsity of the datasets. This issue limits the discriminative power of WTA. We then propose a solution for this problem based on the idea of Densification which provably fixes the issue. Our experiments show that Densified WTA Hashing outperforms Vanilla WTA both in image classification and retrieval tasks consistently and significantly.
\end{abstract}

\section{Introduction}\label{section1}

In many important applications like information retrieval and natural language processing, text documents and images data are in high-dimensional representation. Such high-dimensionality is usually accompanied by extreme data sparsity due to either a large vocabulary or the use of large image window size. The major reason that we find very sparse datasets almost everywhere is because of the wide adoption of the Bag of Words (BoW) representation for documents and images. It is often the case, in BoW representation, that just the presence or absence of specific features carries the most information~\cite{chapelle1999support, jiang2007floating}, especially with higher order shingles. The popularity of sparse codes~\cite{lee2006efficient} for image data is another reason for the abundance of sparse datasets in modern applications. To get a sense of this extreme sparsity, the datasets demonstrated in Google’s Machine Learning system SIBYL~\cite{canini2012sibyl} have dimensions in billions and non-zeros in only few thousands (even hundreds).

With the advent of the internet and the explosion in the volumes of data, almost all machine learning and data mining applications are constrained by their computational requirement. Learning with non-liner kernels, by materializing kernel matrices, which are quadratic in computation and memory, is infeasible~\cite{Proc:Rahimi_NIPS07,li2011hashing,shrivastava2015probabilistic}. Randomized algorithms, especially those based on Locality Sensitive Hashing (LSH)~\cite{indyk1998approximate} have shown huge promise for reducing computational and memory requirement in these scenarios. These randomized algorithms lead to huge gains in computation and memory for a small, insignificant, amount of approximations.


LSH are quite popular for efficient sub-linear algorithms for near neighbor search~\cite{indyk1998approximate}. Even a simple linear scan for near neighbor search, over massive datasets, becomes prohibitively expensive~\cite{Proc:Weber_VLDB98}. There no options but to use hashing approaches for such scenarios. LSH algorithms can also be used as cheap random kernel features~\cite{li2011hashing} for training large-scale non-linear SVMs without materializing the expensive kernel matrix, leading to linear time algorithms. LSH based algorithms are embarrassingly parallel, simple and cheap. Owing to these unique advantages they are heavily used by commercial search industries for truly large-scale data processing systems.

\begin{figure*}[t!]
	\centering
	\renewcommand{\arraystretch}{1.7}
	
	\centering
	\begin{tabular}{ |>{\centering\bf }m{2cm}|>{\centering\bf}m{2.5cm}|>{\centering\bf}m{2.5cm}|>{\centering\bf}m{2.5cm}|>{\centering\bf}m{2.5cm}|m{2cm}}\cline{1-5}
		& $\bf x_1$ & $\bf x_2$ & $\bf x_3$ & $\bf x_4$ &
		\\\cline{1-5}
		$\bf x$ & 10, 12, 9, 23 & 8, 9, 1, 12 & 9, 2, 6, 1 & 3, 5, 1, 7 & \\\cline{1-5}
		$\bf \Theta(x)$ & {\color{red}{23}}, 10, 12 & {\color{red}{12}}, 8, 9 & 1, {\color{red}{9}}, 2 & {\color{red}{7}}, 3, 5 & \\\cline{1-5}
		$\bf \mathcal{H}_{wta}(x)$ & 1 & 1 & 2 & 1 & \\\cline{1-5}
	\end{tabular}
	
	\caption{WTA Hashing Example with four input vectors $x_1, x_2, x_3, x_4$, $K=3$ and one permutation $\Theta = 4, 1, 2$}
	
	\label{fig:1}
\end{figure*}

In the last decade, similarities based on relative (or comparative) attributes have gained huge popularity especially in the vision literature~\cite{parikh2011relative}. For such similarities, a well-known hashing scheme is \emph{Winner Take All (or WTA) hashing}~\cite{yagnik2011power}. It is one of the fastest known hashing scheme, which is much faster than signed random projection (SRP). SRP requires one pass over the data vector for computing one hash value. On the contrary, WTA can generate multiple hashes in one pass. It is widely known that hashing time is the major bottleneck, both in theory and practice, for the task of image retrieval. This is the main reason why Google~\cite{dean2013fast} needed WTA for detecting 100,000 objects on single machine in near-real time with very respectable accuracy. Large-scale image retrieval, with low-latency constraints, is a reality and we cannot afford to have costly hash functions, since even one pass over the data vector for hash computation is prohibitively expensive both for energy and latency. WTA hashing has been quite successfully applied to produce superior results on massive-scale object recognition and information retrieval. This randomized hashing scheme seems quite suitable for taking advantage of multiple partial order statistics rather than total orderings of the input vector's feature dimensions to produce sparse embedding codes.

\vspace*{0.5\baselineskip}

{\bf Our Contributions:} In this work, we study the applicability of WTA hashing for very sparse datasets. We found that WTA hashes are not very informative for sparse datasets. We further provide a remedy based on the recent idea of \emph{Densification}~\cite{shrivastava2014densifying,Proc:Shrivastava_UAI14}. In particular, our contributions can be summarized as follows:
\begin{enumerate}
	\item We illustrate that the popular WTA hashing scheme starts loosing information for very sparse datasets, i.e., most of the hash values for very sparse datasets do not have enough discriminative information.
	\item We provide a solution and propose Densified WTA Hashing which combines traditional WTA hashing~\cite{yagnik2011power} with the idea of Densification~\cite{shrivastava2014densifying,Proc:Shrivastava_UAI14}. We show that the idea of densification provably fixes the issue. Our proposal is simple to implement and requires minimal modifications to the original WTA hashing. Furthermore, for dense datasets, our proposal is equivalent to the original WTA hashes and thus a smooth generalization of WTA for sparse datasets.
	\item Previously the idea of Densification was only known to speedup hash functions without loosing quality. For the first time, we show that the idea of Densification actually leads to significant improvement in the quality of WTA hashing, informative hashes. This can be of independent interest in itself.
	\item We demonstrate the benefits of our proposal by showing significant gains in accuracy compared to WTA on real sparse datasets for both retrieval and classification tasks.
\end{enumerate}




\section{Review WTA Hashing}

~\cite{parikh2011relative} pointed out the importance of relative attributes in the vision community. It suggested that for a given vector $x$, the information that given attribute $x_i$ is dominant over some other attribute $x_j$ has stronger discriminative powers compared to other features. It was further shown in~\cite{yagnik2011power} that comparative reasoning (or order information) among attributes is a very informative feature and similarities based on such comparisons lead to superior performances compared to widely adopted measures like $L_2$ distances. However, kernel based (or similarity based) learning is computationally slow. To mitigate this problem, WTA (Winner Takes ALL) Hashing was proposed~\cite{yagnik2011power}. The simplicity, scalability, and power of WTA hashing were quite appealing and it has been successfully used by commercial big-data companies to scale up the task of object detection significantly~\cite{dean2013fast}.

WTA hashing generates a set of random sample of $K$ attributes, typically using a random permutation $\Theta$, and stores the index of the attribute with the maximum weight. It can be implemented in several lines with matlab:

\vspace*{1\baselineskip}
{\bf function $[$maxval, c$]$ = wta(X,K)

theta = randperm(size(X,2));

$[$maxval, c$]$ = $max$(X(:,theta(1:K)), $[]$, 2); \\}


\subsection{Key WTA Notations}
We denote $\Theta(x)$ to be the $K$ random samples from $x$ sampled using permutation $\Theta$. For convenience, we drop the dependence on $K$ as it will remain a fixed constant. $\mathcal{H}_{wta}(\Theta(x))$ will indicate the corresponding WTA hash value. We will also drop $\Theta$ and use $\mathcal{H}_{wta}(x)$ whenever it is clear.

As illustrated in the example shown in Figure \ref{fig:1}, the original input vectors $x_1, x_2, x_3, x_4$ are applied with random permutation $\Theta = (4, 1, 2, 3)$ and first $K=3$ attributes of the permuted vectors are selected (random sample of size 3), e.g. Vector (a) $ =[10, 12, 9, 23]$ will sample $[23, 10, 12]$. Then the index of the maximum attribute in every transformed vector is stored separately, e.g. 1 for (a), to contribute to the final WTA hash codes.

It was shown that WTA hashing scheme has locality sensitive hashing (LSH) property~\cite{yagnik2011power}. It implies that collision probability under this scheme, i.e. for given vectors $x$ and $y$, $Pr(\mathcal{H}_{wta}(x) = \mathcal{H}_{wta}(y)) = \mathbb{E} [{{\bf I}_{\mathcal{H}_{wta}(x) = \mathcal{H}_{wta}(y)}}]$ is some desirable order based similarity measure. It was later shown that for $K=2$ this similarity is the well known Kendall Tau~\cite{Ziegler2012NIPS}.

\begin{figure*}[!htbp]
	
	\renewcommand{\arraystretch}{1.7}
	\centering
	\begin{tabular}{ m{1.5cm}|>{\centering\bf }m{1.5cm}|>{\centering\bf }m{6cm}|m{0.1cm}}\cline{2-3}
		
		&$\bf x_1$ & 0, 0, 5, 0, 0, 7, 6, 0, 0 & \\\cline{2-3}
		&$\bf x_2$ & 0, 0, 1, 0, 0, 0, 0, 0, 0 & \\\cline{2-3}
	\end{tabular}
	\vspace*{2\baselineskip}\newline
	\centering
	\begin{tabular}{ |>{\centering\bf }m{2.3cm}|>{\centering\bf}m{1.5cm}|>{\centering\bf}m{1.5cm}|>{\centering\bf}m{1.5cm}|>{\centering\bf}m{1.5cm}|>{\centering\bf}m{1.5cm}|>{\centering\bf}m{1.5cm}|m{1.5cm}}\cline{1-7}
		& Sample 1 & Sample 2 & Sample 3 & Sample 4 & Sample 5 & Sample 6 & \\\cline{1-7}
		$\bf \Theta$ & 2, 1, 8 & 5, 3, 9 & 6, 2, 4 & 8, 9, 1 & 1, 7, 3 & 2, 4, 5 & \\\cline{1-7}
		$\bf \Theta(x_1)$ & {\color{red}{0}}, 0, 0 (E)& 0, {\color{red}{5}}, 0 & {\color{red}{7}}, 0, 0 & {\color{red}{0}}, 0, 0 (E)& 0, {\color{red}{6}}, 5 & {\color{red}{0}}, 0, 0 (E)& \\\cline{1-7}
		$\bf \Theta(x_2)$ & {\color{red}{0}}, 0, 0 (E)& 0, {\color{red}{1}}, 0 & {\color{red}{0}}, 0, 0 (E) & {\color{red}{0}}, 0, 0 (E)& 0, 0, {\color{red}{1}} & {\color{red}{0}}, 0, 0 (E)& \\\cline{1-7}
		$\bf \mathcal{H}_{wta}(x_1)$ & 1 (E)& 2 & 1 & 1 (E)& 2 & 1 (E)& \\\cline{1-7}
		$\bf \mathcal{H}_{wta}(x_2)$ & 1 (E)& 2 & 1 (E)& 1 (E)& 3 & 1 (E)& \\\cline{1-7}
	\end{tabular}
	\caption{WTA Hashing with input vectors $x_1, x_2$ and six Samples generated using six permutations. E denoted an empty sampling. WTA treats E and E as match of hash values, which artificially inflates the similarity perceived by the hashes.}
	\label{fig:2}
\end{figure*}
\vspace*{1\baselineskip}

\section{Sparse Datasets and Issues with WTA Hashing}\label{sec3}

WTA hashing and the idea of comparative reasoning is quite appealing and intuitive. In this section, we delve deeper and show a critical issue with WTA hashing. We show that for very sparse datasets, which are common in practice~\cite{li2011hashing}, WTA based hashes are not very informative and deviate from the "relative attribute" intuition. We use the equivalence between hashing and the kernel view to illustrate this issue. With every hashing scheme $\mathcal{H}$ is an associated positive definite kernel given by the collision probability $Pr(\mathcal{H}(x) = \mathcal{H}(y)) = \mathbb{E}[ {{\bf I}_{\mathcal{H}(x) = \mathcal{H}(y)}}]$. For large scale learning, as shown in~\cite{li2011hashing,yagnik2011power}, we can convert these hashes into random kernel features~\cite{Proc:Rahimi_NIPS07} by converting hash values to indicator vectors. Please refer~\cite{li2011hashing,yagnik2011power} for more details.

\subsection{Sparsity makes WTA Uninformative.}
Define the sparsity of a dataset $X$ with $n$ samples, with each sample of dimension $d$, as
\begin{equation}
S_x = \frac{\sum_{i=1}^{n}\sum_{j=1}^{d}[1\{X_{ij}=0\}]}{n\times d}
\end{equation}
Note that $[1\{X_{ij}=0\}]$ is an indicator for the event $X_{ij} =0$. $S_x$ is also the probability that $Pr(X_{ij}=0)$.
We will show that the kernel associated with WTA hashing becomes uninformative as the sparsity of dataset increases.


Consider the example that is shown in Figure~\ref{fig:2}. We have very sparse input vectors $x_1$, $x_2$ and we generate six WTA hashes with $K=3$. In order to do this, we sample $K=3$ attributes six different times. Each different sample is generated using six different permutations. Due to sparsity many of these samples are all zeros. We can see that in all the samples except Sample 5, $\mathcal{H}_{wta}((x_1))$ and $\mathcal{H}_{wta}((x_2))$ collide and therefore the estimated collision probability, from the hashes, is roughly $\frac{5}{6}$ indicating high similarity (1 is maximum). This seems misleading.

Due to sparsity, it is very likely that for a given $x$, all the sampled attributes $\Theta(x)$ are zeros for some samples. We represent this situation by $\Theta(x) =E$ (Empty).
Consider Sample 1, 4 and 6, they collide only because they are all zeros. Note, WTA treats all empty samples as collisions and two empty samples will always lead to a hash collision. Sparse datasets are common with Bag-of-Words (or token based) representation. Given $x_1$ and $x_2$, empty Samples (1, 4 and 6) indicate the absence of the randomly chosen $K$ tokens. This is not a strong indicator of similarity. In BoW analogy, if two documents concurrently lack the words ''Hashing", ''Winner" and ''Take", it does not indicate strong similarity given the large vocabulary and the sparse nature of the dataset. In sparse BoW representation, the absence of features is not informative. Only presence of features is important. Thus, whenever both the $K$ samples, under considerations for WTA, are empty we observe undesirable collision. Note, that if we treat them as mismatch then it is also a problem. In hashing if hashes do not collide it is an indicator that the points are not similar. Making empty samples not collide will treat sparsity as dissimilarity, which is again undesirable. Thus, there is no straightforward fix to this problem. 

\vspace*{1\baselineskip}
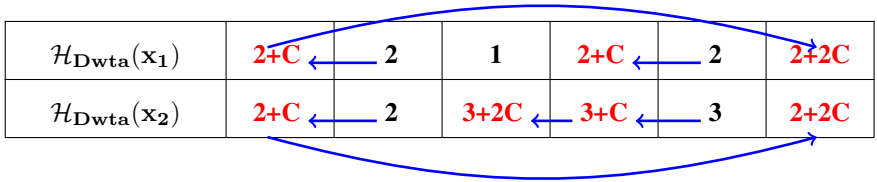
\begin{figure*}[!htbp]
	\centering
	\renewcommand{\arraystretch}{1.7}
	\begin{tabular}{ |>{\centering\bf}m{2.5cm}|>{\centering\bf}m{1cm}|>{\centering\bf}m{1cm}|>{\centering\bf}m{1cm}|>{\centering\bf}m{1cm}|>{\centering\bf}m{1cm}|>{\centering\bf}m{1cm}|m{1cm}}\cline{1-7}
		$\bf \mathcal{H}_{Dwta}(x_1)$ & {\color{red}{2\tikzmark{g}+C}} \tikzmark{a} & \tikzmark{b} 2 & 1 & {\color{red}{2+C}} \tikzmark{e} & \tikzmark{f} 2 & {\color{red} {2+\tikzmark{c}2C}} & \\\cline{1-7}
		$\bf \mathcal{H}_{Dwta}(x_2)$ & {\color{red}{2\tikzmark{g1}+C}} \tikzmark{a1} & \tikzmark{b1} 2 & {\color{red}{3+2C}} \tikzmark{h1} & \tikzmark{i1} {\color{red}{3+C}} \tikzmark{e1}& \tikzmark{f1} 3 & {\color{red}{2+\tikzmark{c1}2C}} & \\\cline{1-7}
	\end{tabular}
	\begin{tikzpicture}[overlay, remember picture, yshift=.25\baselineskip, shorten >=.5pt, shorten <=.5pt]
	\draw [->] [line width=1pt, color=blue] ({pic cs:b})-- ({pic cs:a});
	\draw [->] [line width=1pt, color=blue] ({pic cs:f})-- ({pic cs:e});
	\draw [->] [line width=1pt, color=blue] ([yshift=5.75]{pic cs:g}) [bend left=15] to ([yshift=5.75]{pic cs:c});
	\draw [->] [line width=1pt, color=blue] ({pic cs:b1})-- ({pic cs:a1});
	\draw [->] [line width=1pt, color=blue] ({pic cs:f1})-- ({pic cs:e1});
	\draw [->] [line width=1pt, color=blue] ({pic cs:i1})-- ({pic cs:h1});
	\draw [->] [line width=1pt, color=blue] ([yshift=-6]{pic cs:g1}) [bend right=15] to ([yshift=-6]{pic cs:c1});
	\end{tikzpicture}
	\vspace*{2\baselineskip}
	\caption{Example densification of WTA hashes shown in Figure~\ref{fig:2}. All the empty bins are reassigned (shown in red) by borrowing values from non-empty samples (shown by arrow). This unusual procedure actually is the right fix for WTA as shown by Theorem~\ref{theo:main}}
	\label{fig:3}
\end{figure*}

If we further observe Sample 3, the collision is even worse because it is meaningless that an empty Sample of $x_2$ collides with a non-empty sample of $x_1$, simply because the max value in $x_1$ happens to be at index 1. This is actually a spurious collision and can be easily eliminated if we assign special values to all empty samples. So we ignore this, easily fixable but spurious collision, from analysis.

In Sample 2, neither $\Theta(x_1)$ nor $\Theta(x_2)$ are E, so those are informative collisions. It clearly indicates that among the chosen $K$ attributes, in both $x_1$ and $x_2$, the same feature dominates. This is in line with the original motivation of WTA. Owing to the presence of empty samples, although $x_1, x_2$ show high similarity according to WTA Hashing, their sparsities contribute a large fraction of their similarity. This is not quite desirable.

Formally, given vectors $x_1$, $x_2$ and a permutation $\Theta$, define the indicator vector for empty sampling of both $x_1$ and $x_2$:

\begin{equation}
\label{eq:5}
I_{empty}=\left\{
\begin{array}{c l}
1 & \Theta(x_{1})= \Theta(x_{2})=E\\
0 & otherwise
\end{array}\right.
\end{equation}

Note if any of the $\Theta(x_{1})$ is not empty then $I_{empty}=0$. Based on this indicator variable, we can define empty and non-empty collisions as:

\begin{align}
k_{bad}(x_1,x_2) &= Pr(\mathcal{H}_{wta}(x_{1}) = \mathcal{H}_{wta}(x_{2}) | I_{empty}=1) \\
k_{good}(x_1,x_2) &= Pr(\mathcal{H}_{wta}(x_{1}) = \mathcal{H}_{wta}(x_{2}) | I_{empty}=0)
\end{align}

As argued, $k_{bad}(x_1, x_2)$ is not an informative kernel for very sparse datasets. Using these quantities we can formally write the WTA kernel as

\begin{align}
k_{wta}(x_1, x_2) &= Pr(\mathcal{H}_{wta}(x_{1})=\mathcal{H}_{wta}(x_{2})) \\
&= ak_{bad}(x_1, x_2) + (1-a)k_{good}(x_1, x_2),
\end{align}

where $a$ is the probability of $I_{empty}=1$. Clearly, for very sparse datasets $a$ will be high and hence $k_{bad}(x_1,x_2)$ dominates the WTA kernel making it less discriminative.

\section{Our Proposal: Densified WTA Hashing}
\label{others}

In~\cite{shrivastava2014densifying,Proc:Shrivastava_UAI14} the authors proposed the idea of Densification of hashes for obtaining a one-pass hashing scheme which has the same collision probability as the traditional
minwise hashing. The idea was to reassign empty samples, having all zero values, by borrowing values from nearest non-empty samples added with some constant offset.
Motivated by this idea, we propose a similar reassignment of empty samples generated from WTA. We will show that the modified WTA, which we call "Densified WTA" (DWTA) hashing, produces the right kernel. This is little surprising because \emph{Densification} was used in the literature to speed up the hashing scheme with the same old property. Here we rather show a first example where densification improves the hashing scheme by making it more informative.

Vanilla WTA assigns all empty samples a constant value of $1$. Using densification, we assign new random values to all the empty samples. For a given data $x$, we first generate a set of WTA hashes and place them one after the other (See Figure~\ref{fig:3}). Then for all the empty samples we locate the closest non-empty sample towards the cyclic right. Then the newly assigned value to the empty sample is the value of this closest non-empty sample with some appropriate offset. The offset is simply the (cyclic) distance of the closest bin multiplied by any constant $C > K$.

The overall procedure of Densification for reassigning the empty samples from Figure~\ref{fig:2} is shown in Figure~\ref{fig:3}. For $x_1$, Sample 1 and 6 are reassigned with new hash values $2+C$, $2+2C$ and 2 is the base hash value which they borrowed from Sample 2. For Sample 1, it is obvious that Sample 2 is its right closest non-empty sample. But since Sample 6 is the last sample which does not have right non-empty samples, we need to suppose that all the samples lie in a circle so Sample 2 would be the its first right non-empty sample. Moreover Sample 2 is 2 samples away from Sample 6 and therefore $2+2C$ would be the new hash value of Sample 6. Similarly Sample 4 is assigned with $2+C$. Reassignments with the same manner happen to $x_2$ but since it is more sparse than $x_1$, more samples are filled with new hash values.

\begin{figure*}[!htbp]
	
	\renewcommand{\arraystretch}{1.7}
	\centering
	\begin{tabular}{ |>{\centering\bf }m{3cm}|>{\centering\bf}m{1.5cm}|>{\centering\bf}m{2.0cm}|>{\centering\bf}m{1.5cm}|>{\centering\bf}m{2.0cm}|>{\centering\bf}m{1.5cm}|>{\centering\bf}m{2.0cm}|m{1.7cm}}\cline{1-7}
        & \multicolumn{2}{c|}{\bf 1000 BoW ($\%$)}  & \multicolumn{2}{c|}{\bf 5000 BoW ($\%$)} & \multicolumn{2}{|c|}{\bf 10000 BoW ($\%$)}& \\\cline{1-7}
       	&Raw Data Sparsity & {\bf Empty Hash Codes (ratio)} &Raw Data Sparsity &Empty Hash Codes (ratio)&Raw Data Sparsity &Empty Hash Codes (ratio) & \\\cline{1-7}
		\bf VOC2010 &$68.63$ & $23.84$ & $88.18$  & $61.39$  & $92.87$  &  $74.81$ & \\\cline{1-7}
		\bf LabelMe-12-50k &$58.07$ & $13.63$ & $82.93$ & $48.18$  & $89.49$  &  $64.43$ & \\\cline{1-7}
		\bf MSRc &$69.46$ & $24.66$ & $86.83$  & $56.60$  & $91.54$  &  $70.07$ & \\\cline{1-7}
	\end{tabular}
	\caption{Each entry displays the Sparsity of VOC2010, LabelMe-12-50k and MSRc datasets in 1000 BoW, 5000 BoW and 10000 BoW representation. Raw Data Sparsity shows the sparsity of original BoW vectors and ratio of Empty Hash Codes shows the ratio of empty codes in resulting WTA Hashing encoding (empty codes means empty sampling). By increasing dictionary size, Sparsity naturally goes up in all three datasets. }
	\label{fig:sparse}
\end{figure*}

Recall in Section~\ref{sec3} we discuss that the collisions between $\mathcal{H}_{wta}(x_1)$ and $\mathcal{H}_{wta}(x_2)$ happened in Sample 1, 4 and 6. After densification, there is no collision in Sample 4. Therefore after densification the hash collision similarity comes down to $\frac{3}{6} = 0.5$.

Formally, let us assume that we want to generate $n$ hash values. $\Theta_i(x)$ denote sample $i$. Let $CN(i)$ be the smallest number greater than $i$ such that $\Theta_{CN(i) \ mod (n+1) + 1} \ne E$, i.e. the closest non-empty sample towards circular right. We can define the Densified WTA, $\mathcal{H}_{Dwta}$, as follows
\begin{multline}\label{eq:DWTA}
{}\mathcal{H}_{Dwta}(\Theta_i(x))\\
=\left\{
\begin{array}{c l}
\mathcal{H}_{wta}(\Theta_i(x)) & \text{if} \ \ \Theta_i(x) \ne E\\
\mathcal{H}_{wta}(\Theta_{CN(i)}(x)) + C(CN(i) - i) & otherwise
\end{array}\right.
\end{multline}

Based on this definition, we now show our main result that $\mathcal{H}_{wta}$ precisely fixes the issue of non-empty bin and get rid of the bad kernels. Since the result holdS for any sample, we will drop the subscript $i$. Formally,
\begin{theorem}
	\label{theo:main}
	For any given $x$ and $y$, the collision probability of "Densified WTA" $\mathcal{H}_{Dwta}$ satisfies:
	\begin{align}\notag
	Pr(\mathcal{H}_{Dwta}(x_{1}) &= \mathcal{H}_{Dwta}(x_{2})) \\
	&= k_{good}(x_1, x_2) = k_{Dwta}(x_1, x_2),
	\end{align}
\end{theorem}
{\bf Proof:} The proof is a simple case based analysis. Note that $C$ is always greater than the value of any non empty bin and define the index of the maximum attribute of vector
\vspace*{0.5\baselineskip}
 $x$ as $IndMax(x)$.\\
{\bf Case I:} ($I_{empty}^i =0$)\\
Without loss of generality, let ${\Theta(x_1)}_i \ne E$.  If  we have ${\Theta_i(x_2)} \ne E$, then both of these values are untouched and we get
\begin{equation}
\begin{split}
\label{eqimportant}
&\mathcal{H}_{Dwta}(x_1) = \mathcal{H}_{Dwta}(x_2) \\
&\iff IndMax({\Theta_i(x_1)}) = IndMax({\Theta_i(x_2)}).
\end{split}
\end{equation}
In case if ${\Theta_i(x_2)} = E$, then by the choice of C ,
\begin{equation} \label{eq1}
\mathcal{H}_{Dwta}(x_2) > C > IndMax({\Theta_i(x_1)}) = \mathcal{H}_{Dwta}(x_1).
 \end{equation}
Therefore, either way we have Equation~\ref{eqimportant}  and
\begin{equation}
\begin{split}
&Pr(\mathcal{H}_{Dwta}(x_1) = \mathcal{H}_{Dwta}(x_2)) | I_{empty} = 0 ) \\
&= Pr(IndMax({\Theta_i(x_1)}) = IndMax({\Theta_i(x_2)})) | I_{empty} = 0 ) \\
&= k_{good}(x1, x2)
\end{split}
\end{equation}
{\bf Case II:} ($I_{empty}^i =1$) \\
Let
\begin{equation} \label{eq1}
CN(i)_1 = \min x \hspace{0.1in}\text{s.t} \hspace{0.1in}\Theta_{CN(i)_1 \text{mod }(n+1)+1} \ne E.$$ $$CN(i)_2 = \min x \hspace{0.1in}\text{s.t}
\hspace{0.1in}\Theta_{CN(i)_2 \text{mod }(n+1)+1} \ne E.
\end{equation}
\begin{equation} \label{eq1}
m = \min(CN(i)_1, CN(i)_2),
t = m \text{mod }(n+1)+1.
\end{equation}
\vspace*{5mm}
The definition of $CN(i)_1$ and $CN(i)_2$ implies $I_{empty}^t = 0$.
{\bf Subcase I:} ($CN(i)_1  =  CN(i)_2 = m$)\\
We have
\begin{equation} \label{eq1}
\begin{split}
&\mathcal{H}_{Dwta}(x_1) = IndMax({\Theta_t(x_1)}) + mC \hspace{0.15in}  \text{and}\\ &\mathcal{H}_{Dwta}(x_2) = IndMax({\Theta_t(x_2)}) + mC.
\end{split}
\end{equation}
\vspace*{2mm}
and Equation~\ref{eqimportant}.\\
{\bf Subcase II:} ($CN(i)_1 \ne  CN(i)_2 $)\\
Without loss of generality $CN(i)_1 > CN(i)_2  = m$. Clearly, by definition of $t$, $CN(i)_1$ and $CN(i)_2$, we have
\begin{equation} \label{eq1}
 {\Theta(x_1)}_i = E; \hspace{0.15in}  \text{and}\hspace{0.15in} {\Theta(x_2)}_i \ne E.
\end{equation}
Also,
\begin{equation} \label{eq1}
\begin{split}
\mathcal{H}_{Dwta}(x_1) &= IndMax(\Theta_{CN(i)_1 \text{mod }(n+1)+1}) + CN(i)_1C \\
&>\Theta_{CN(i)_2 \text{mod }(n+1)+1} + CN(I)_2C \\
&= \mathcal{H}_{Dwta}(x_2).
\end{split}
\end{equation}

Thus, from the two subcases we can write,

\begin{equation} \label{eq1}
\begin{split}
& Pr(\mathcal{H}_{Dwta}(x_1) = \mathcal{H}_{Dwta}(x_2)) | I_{empty} = 1 ) \\
&= Pr(IndMax({\Theta_i(x_1)}) = IndMax({\Theta_i(x_2)})) | I_{empty} = 0 ) \\
&= k_{good}(x1, x2)
\end{split}
\end{equation}

From Theorem~\ref{theo:main}, it is clear that the new kernel is precisely the good kernel $k_{good}(x_1,x_2)$ with no contribution of $k_{bad}(x_1,x_2)$ in $k_{Dwta}(x_1, x_2)$, irrespective of the sparsity.

\subsection{Cost of Densification}

We can see that we incur negligible cost of densification which can be achieved in two lookups over the generated WTA hashes. We will show that this negligible cost leads to huge performance gains in practice. This we believe is one of the many examples where a careful analysis and some mathematics goes a long way in designing simple and significantly better algorithms.

\subsection{Dealing with Large Hash Values}
\label{mod}
It can be seen from Equation~\ref{eq:DWTA} that the value of $\mathcal{H}_{Dwta}(\Theta_i(x))$ can become large due to the term $C(CN(i) - i)$. It turns out that this is not a problem. There is a significant amount of literature to reduce the final range of hashing scheme~\cite{Article:Li_Konig_CACM11}. The idea is to randomly shrink the range at insignificant cost of small constant random collisions. We found that if we want to constrain the final hash value to a range $R$ simply taking mod $R$ of the final hash value suffices in practice. This is also what we use during evaluations.

\begin{figure*}[!htbp]
	\centering
	\mbox{\hspace{-0.15in}
		\includegraphics[width=2.5in]{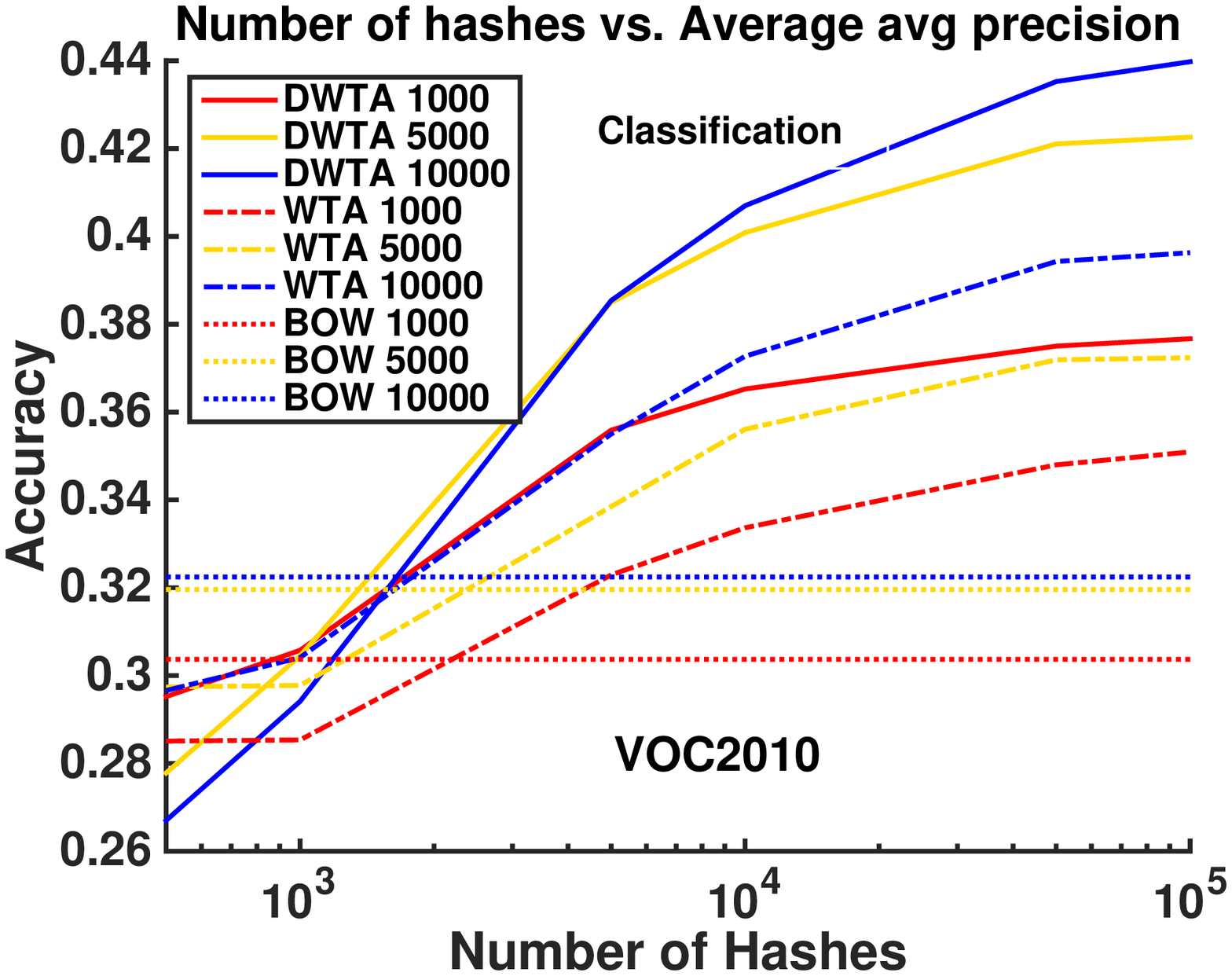}\hspace{-0.2in}
		\includegraphics[width=2.5in]{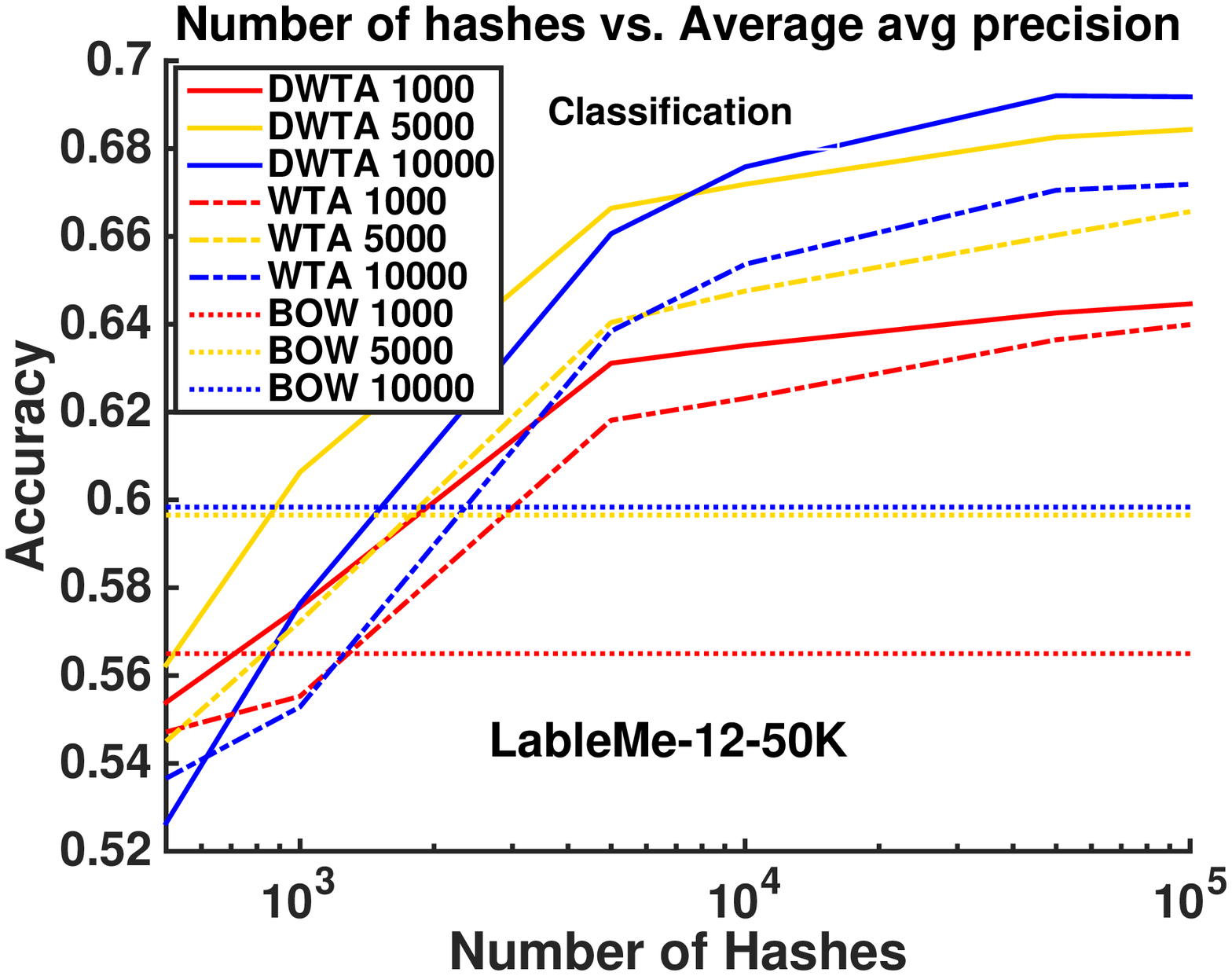}\hspace{-0.2in}
		\includegraphics[width=2.5in]{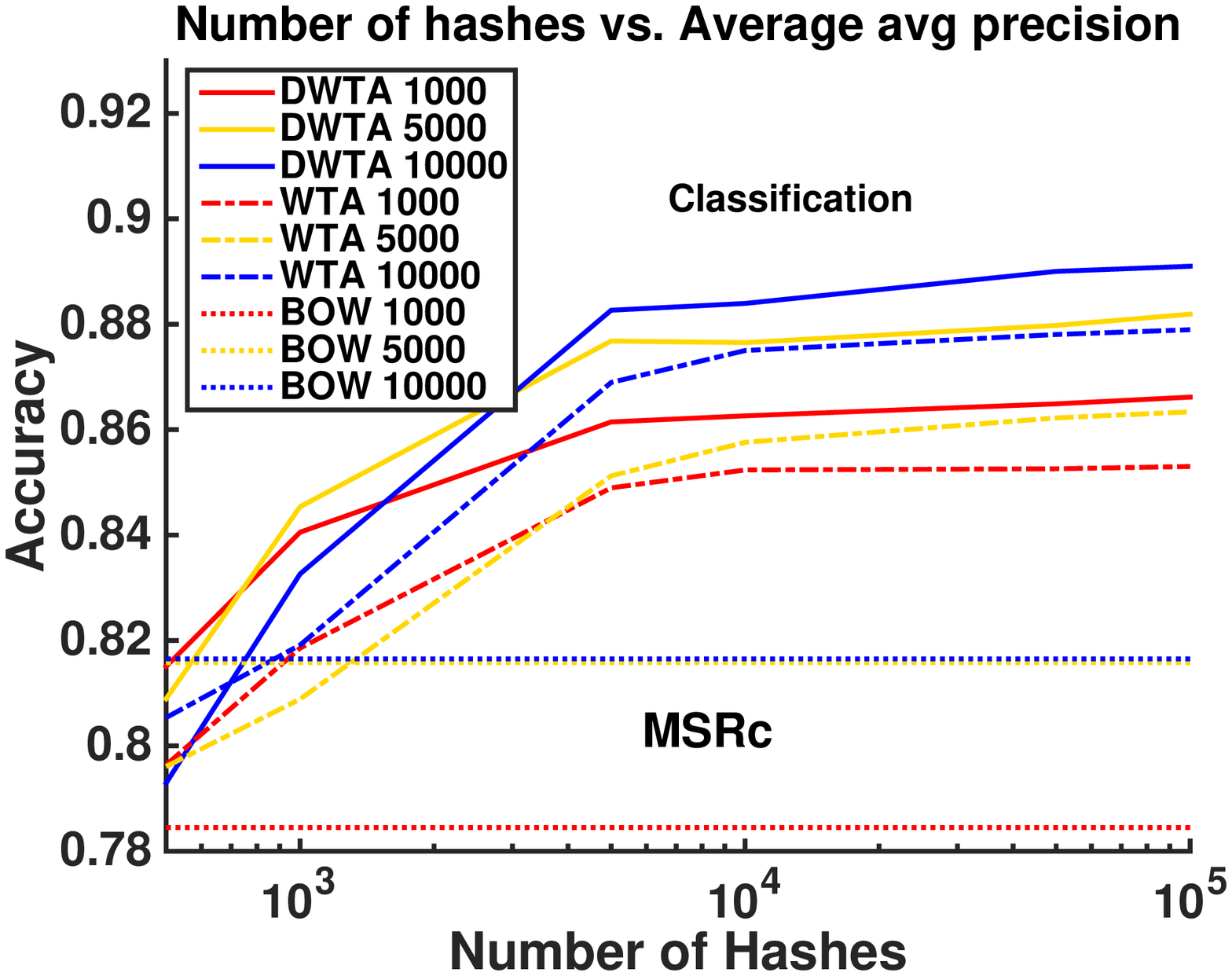}
	}
	
	\vspace{0.1in}
	
	\caption{Densified WTA vs. WTA on the task of Image Classification on three different vision datasets. We used 1000, 5000 and 10000 BoW representation of the images. The y-axis is the mean accuracy and the x-axis is the number of hashes used as features. The horizonal lines (dotted) are classification based just on the BoW features. The semi-dotted lines is the vanilla WTA hashes and bold lines are our proposed Densified WTA Hashes. The colors represent which BoW (among 1000, 5000 and 10000) was used as features. Densified WTA significantly outperforms the corresponding WTA consistently for all the choices.}
	\label{fig:class}
\end{figure*}

\begin{figure*}[!htbp]
	\vspace{0.1in}
	\centering
	\mbox{\hspace{-0.15in}
		\includegraphics[width=2.5in]{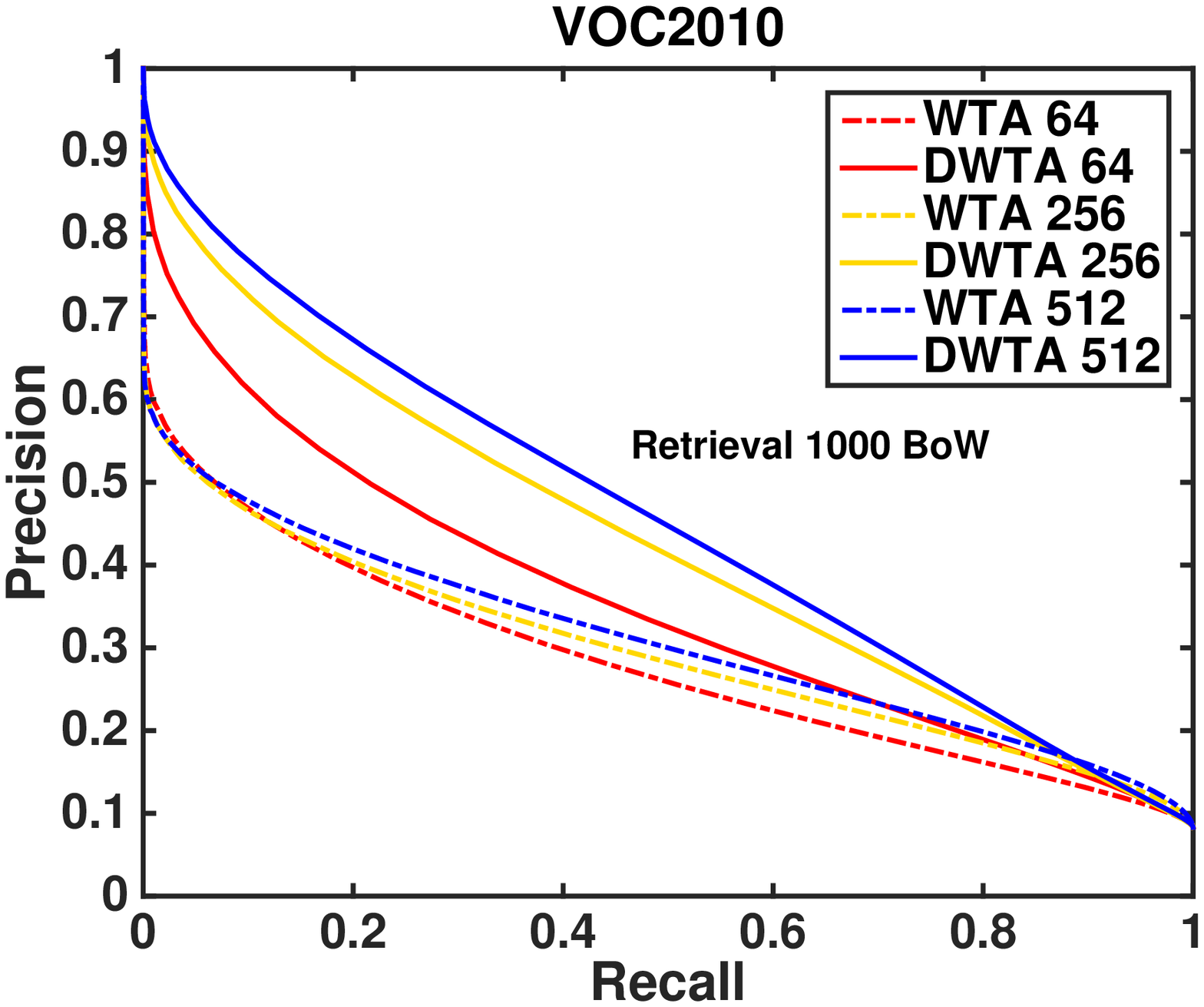}\hspace{-0.2in}
		\includegraphics[width=2.5in]{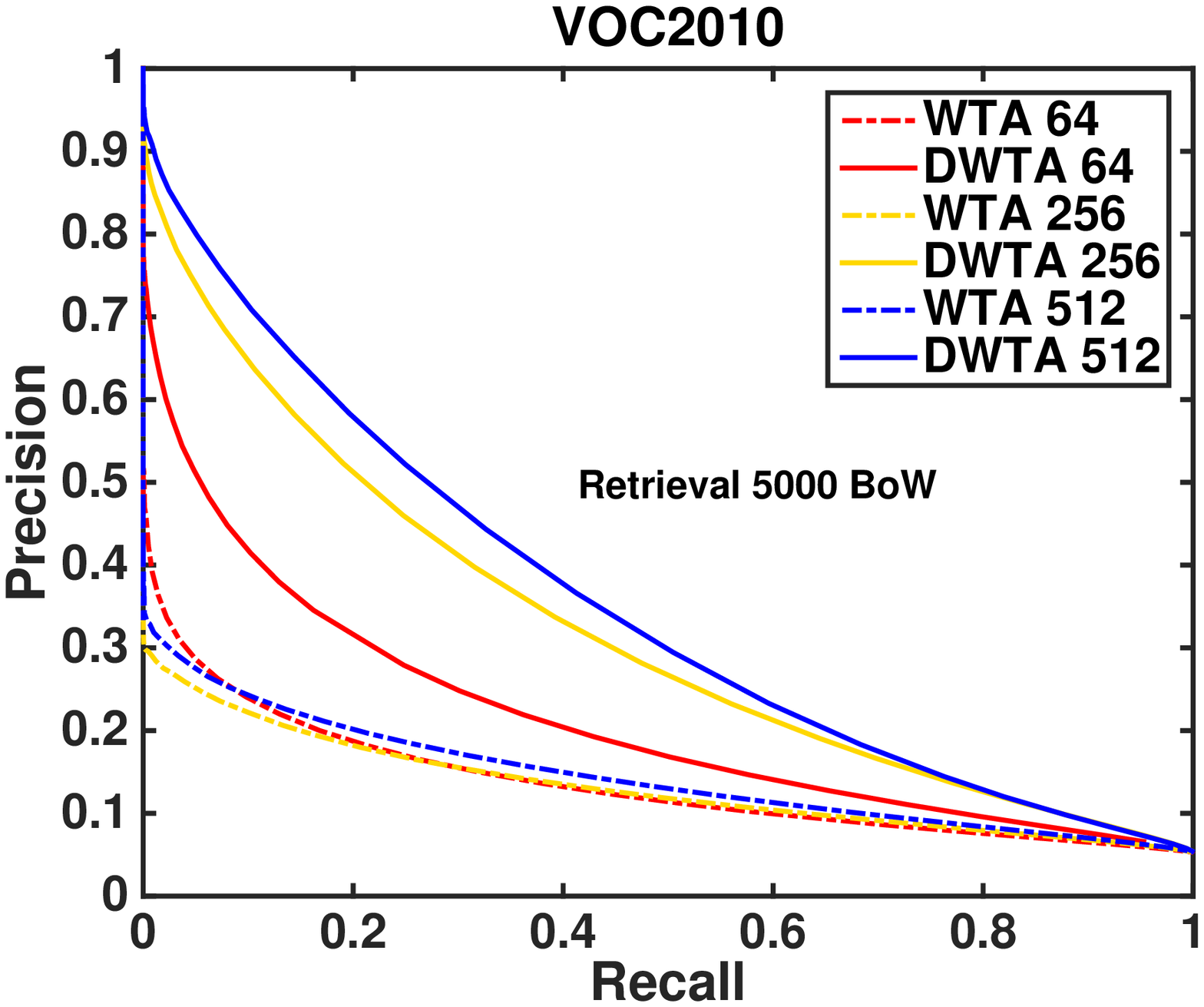}\hspace{-0.2in}
		\includegraphics[width=2.5in]{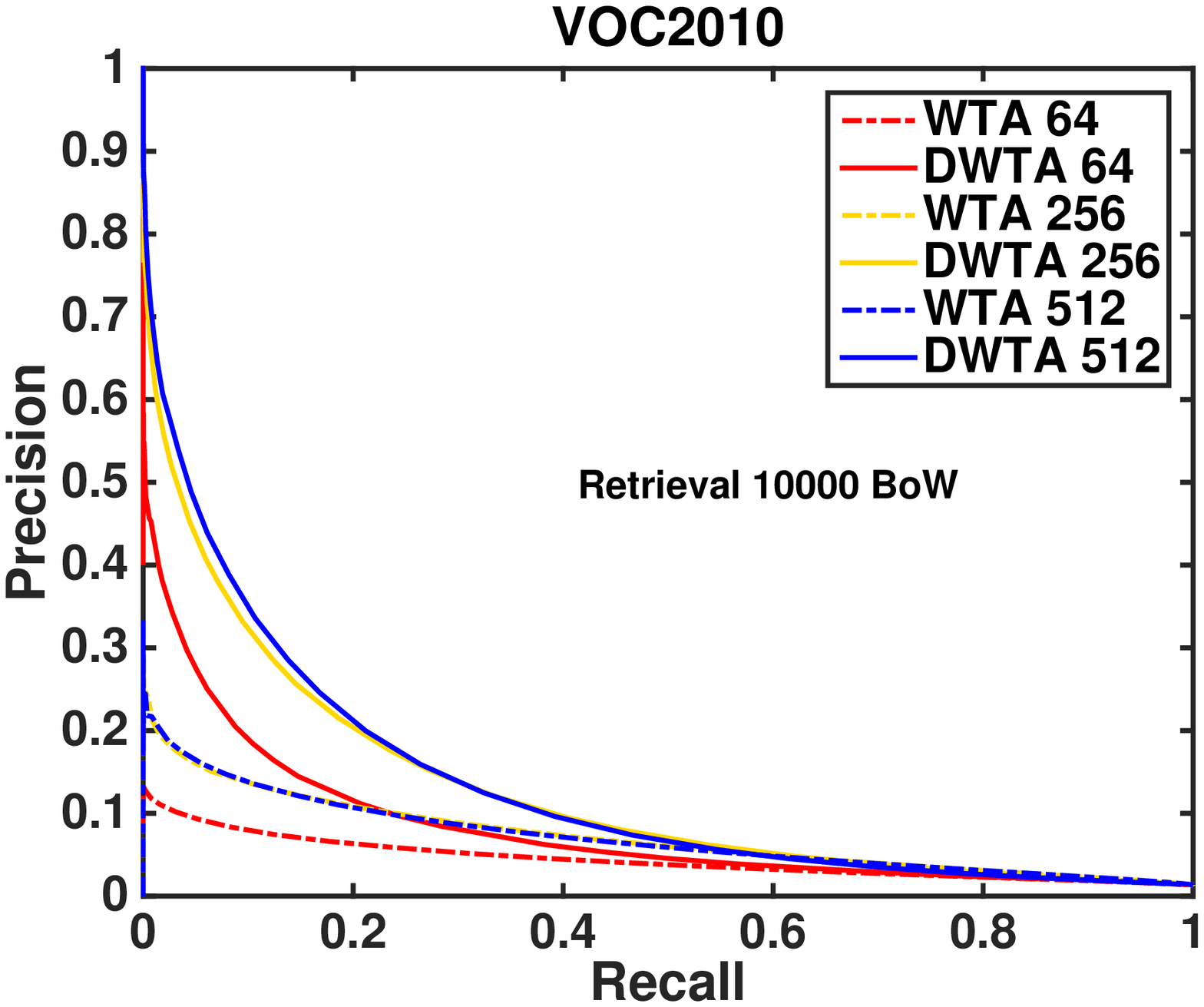}
	}
	
	\vspace{0.1in}
	
	\centering
	\mbox{\hspace{-0.15in}
		\includegraphics[width=2.5in]{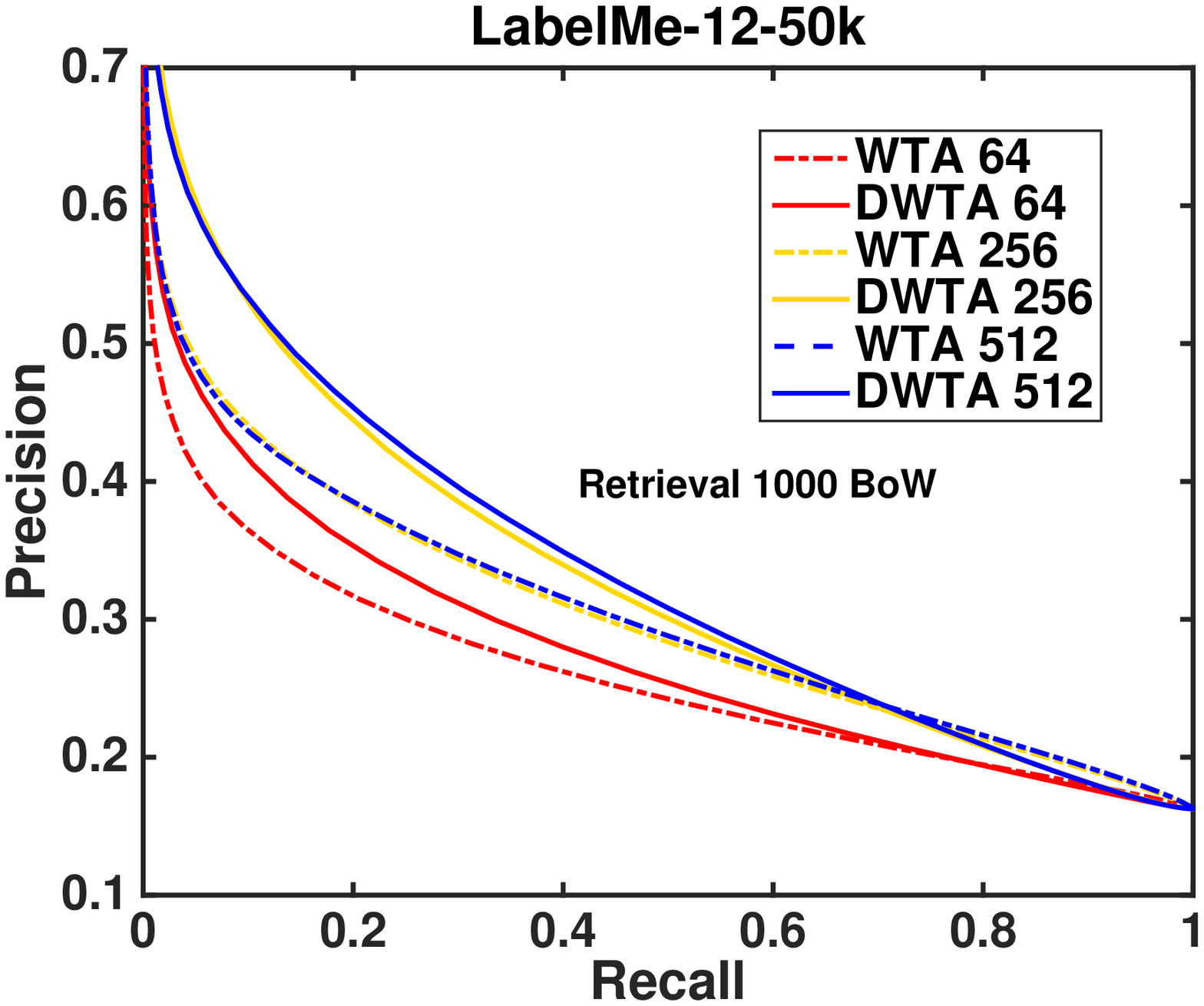}\hspace{-0.2in}
		\includegraphics[width=2.5in]{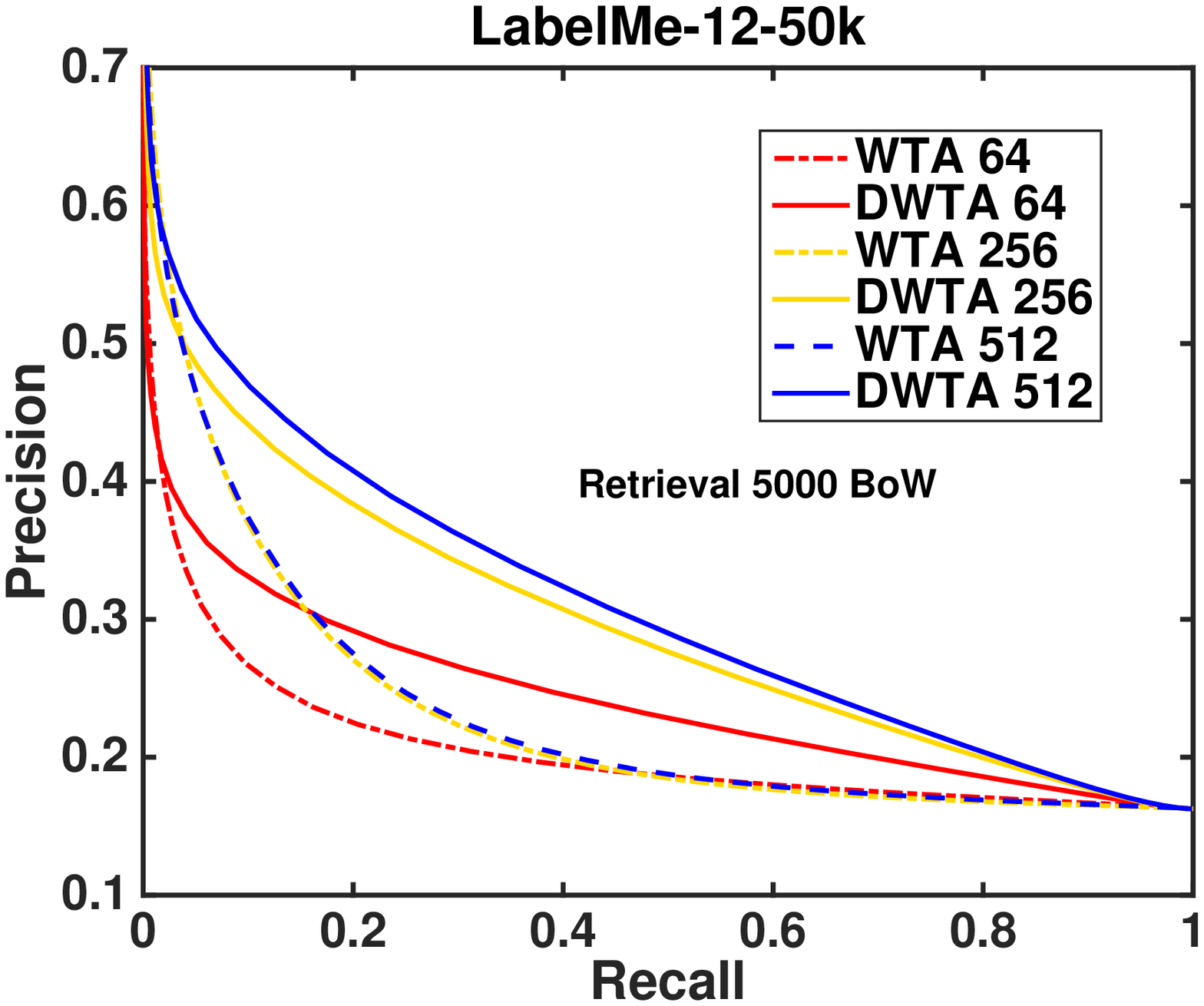}\hspace{-0.2in}
		\includegraphics[width=2.5in]{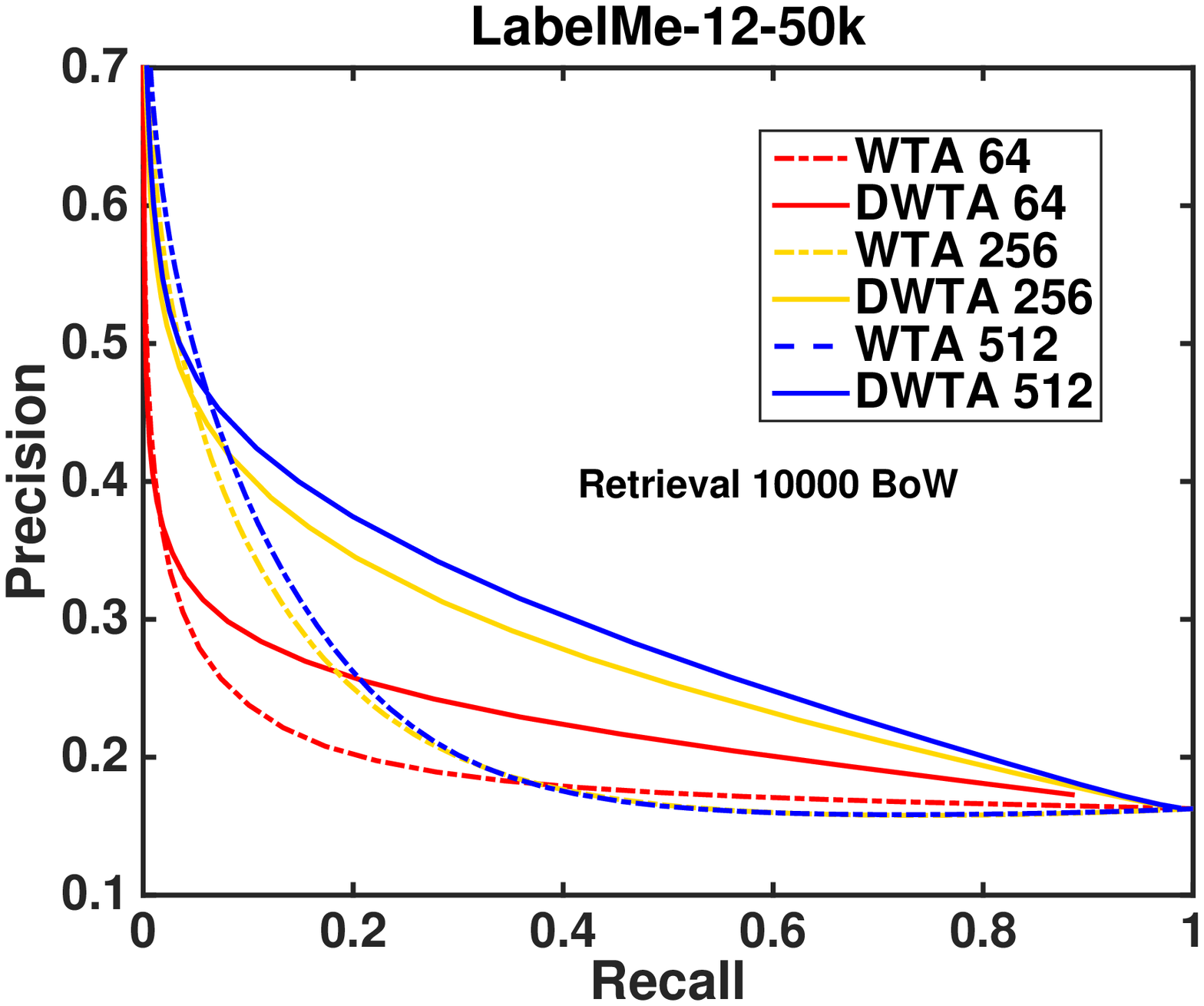}
	}
	
	\vspace{0.1in}
	
	\centering
	\mbox{\hspace{-0.15in}
		\includegraphics[width=2.5in]{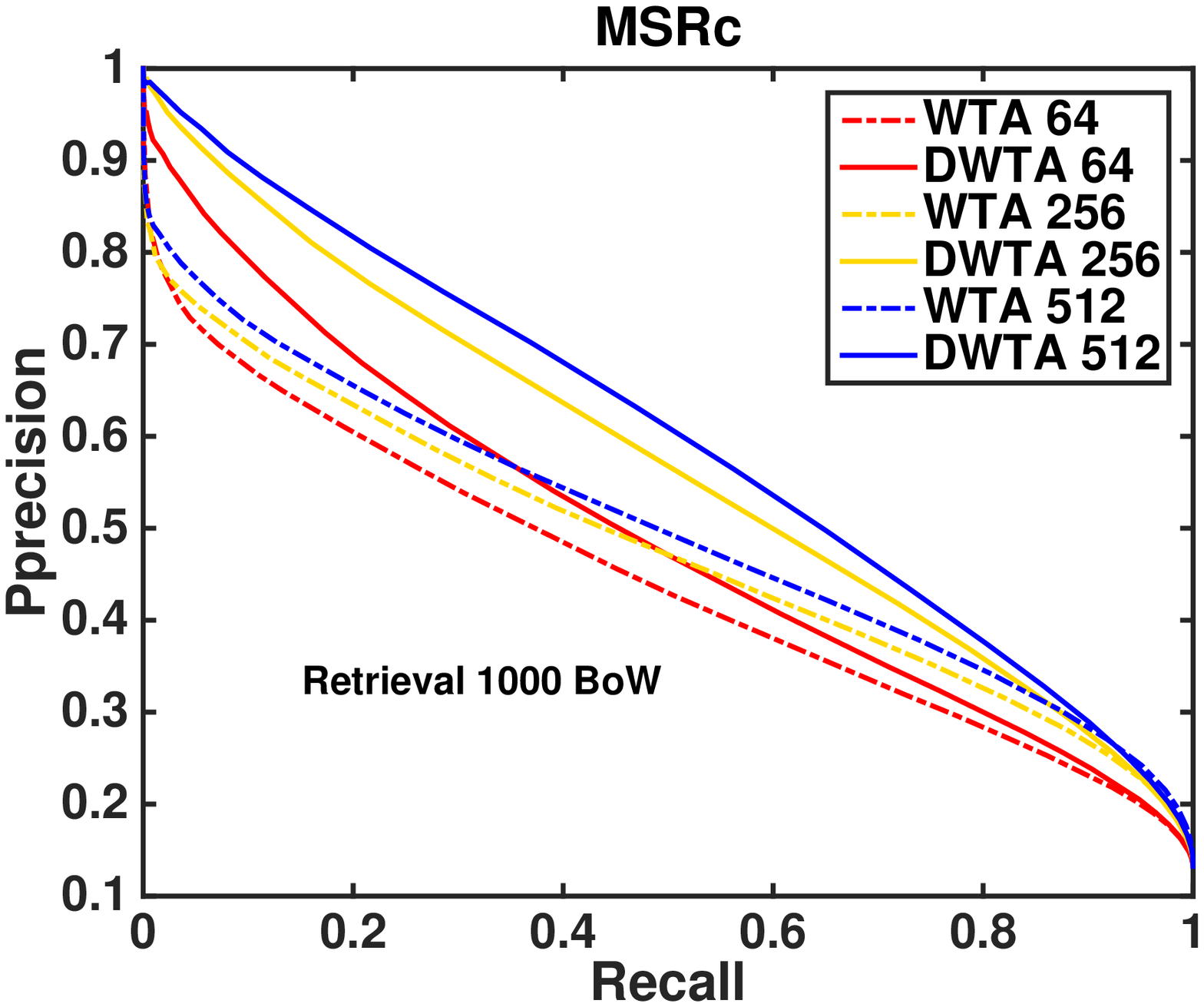}\hspace{-0.2in}
		\includegraphics[width=2.5in]{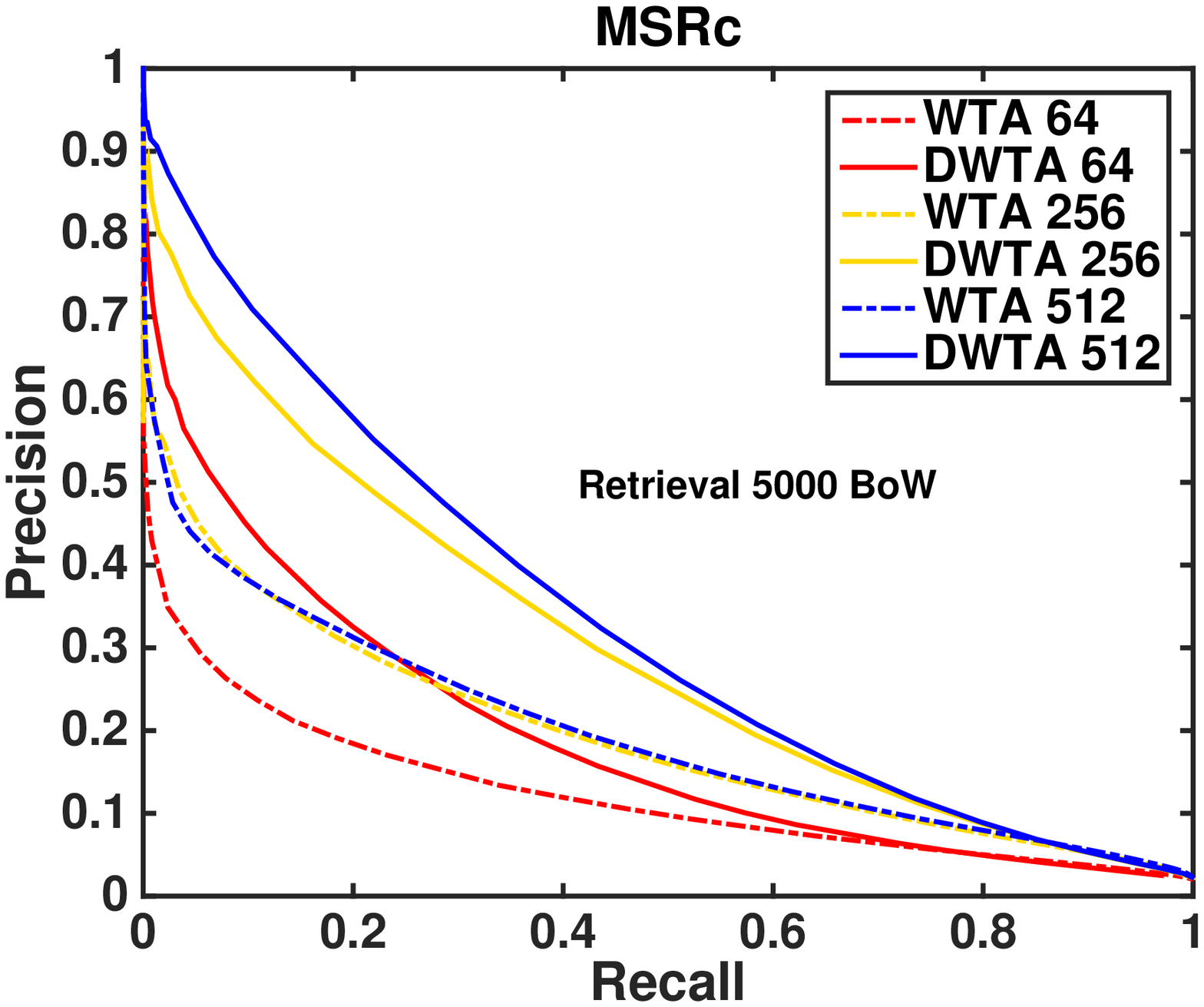}\hspace{-0.2in}
		\includegraphics[width=2.5in]{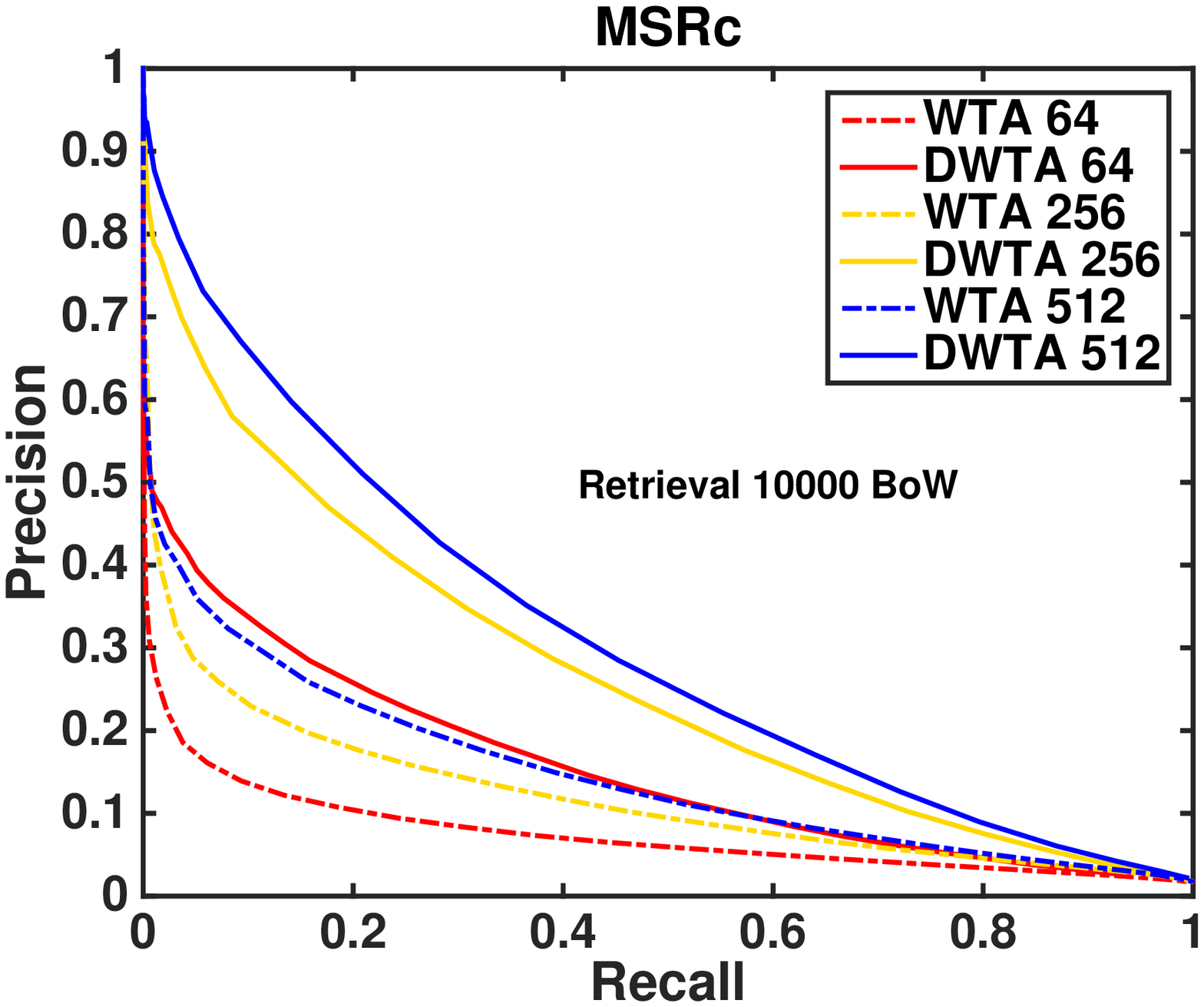}
	}

	\caption{Precision and Recall curves comparing the retrieval performance of Densified WTA vs. WTA on VOC2010, LabelMe-12-50k and MSRc datasets for 1000, 5000 and 10000 BoW feature representations. The semi-dotted lines is the vanilla WTA hashes and bold lines are our proposed Densified WTA Hashes. Different colors represent different number of hashes used for ranking (among 64, 256 and 512). Densified WTA significantly outperforms the corresponding WTA consistently.}
	\label{fig:ret}
\end{figure*}

\section{Experiments}
\label{headings}

In this section, we compare the performance of Densified WTA hashing with Vanilla WTA on two tasks: 1) classification and 2) Image retrieval. They are important task of evaluating the performance of Hashing algorithms, because the classification accuracy can quantify the discriminative power in hashes and hashing has received
increasing interests in efficient large scale image retrieval With the rapid growth of web images.

\subsection{Data sets and baselines}
We use three popular publicly available image data sets, including VOC2010 ~\cite{everingham2010pascal}, LabelMe-12-50k ~\cite{russell2008labelme} and MSRc ~\cite{msrc}:
\begin{itemize}
	\item The VOC2010 database contains a total of 10103 annotated images of twenty classes, including people, animals, vehicles and indoors. The data has been split into 50\% for training and 50\% for testing. One image could belong to different classes.
	\item The LabelMe-12-50k dataset consists of 50,000 JPEG images of twelve classes, 80\% for training and 20\% for testing. They are $256\times256$-pixels pictures extracted from LabelMe~\cite{russell2008labelme}.
	\item The MSRc is a database of thousands of labeled, high-resolution (680x480 pixels) images of eighteen classes.
\end{itemize}
The authors of WTA paper used LabelMe for retrieval task and VOC2010 datasets for classification task. We demonstrate both retreival and classification on both of the datasets as well as a new MSRc dataset. As described in Section~\ref{mod}, to reduce the space of Densified WTA Hashing we apply $mod$ operation on hash values of all samples as a fix.

Table~\ref{fig:sparse} summarizes the sparsity of Raw Data, input Bag of Words, and the ratio of Empty Hash codes, the resulting codes after applying WTA Hashing to input Bag of Words vectors. We can see that when number of BoW increases, sparsity, highest in 10000 BoW, also goes up in all three datasets. Note here, we are doing the same tasks as WTA paper, but we do not apply exactly same settings and the sparsity would thereby be different (they did not reveal sparsity of their datasets as well). Therefore, we do not expect exactly same results on VOC2010 dataset due to the sparsity difference.

%
%

\subsection{Classification }
Our motivation of comparing two Hashing algorithms using classification accuracy is to quantify the discriminative power in hashes. We use Densified WTA codes to do the classification task on the VOC2010, LabelMe-12-50k and MSRc data sets. We don't compare with those state-of-the-art methods like a particular type of nonlinear Mercer kernels, e.g. the intersection kernel or the Chi-square kernel ~\cite{yang2009linear} in classifying these data sets. Instead, we apply Densified WTA and original WTA hashes to a baseline method, sparse BoW of local descriptors and passing to linear SVM classifier, to show that the Densified WTA achieve superior improvement on classification tasks on sparse data.

Replicating the setting of the original WTA paper, we first generated standard BoW of local descriptors, computed from the images, using the publicly available code~\cite{vedaldi2010vlfeat}. BoW was generated by extracting local descriptors from dense grid over each image and quantizing them using K-means. We used DSIFT~\cite{kokkinos2012dense} as our descriptor measuring gradient at each key point pixel. The gradient was represented by a single 128-dimensional vector, stacked by a three-dimensional ($8\times4\times4$) elementary feature vector formed by the pixel location ($4\times4$) and the gradient orientation (8). In this experiment, we consider BoW with 1000, 5000, and 10,000 bins to demonstrate the effect of sparsity.

We then generated WTA and Densified WTA hashes from these images and generates feature vectors as suggested in the WTA paper. For generating WTA and Densified WTA features, we used the fixed recommended setting of $K=4$ for all the datasets. We varied the number of hash features over a range of values: $5\times10^2, 1\times10^3, 5\times10^3, 1\times10^4, 5\times10^4, 1\times10^5$.

To compute the classification performance we ran a simple SVM on BoW features, WTA hashed features, and Densified WTA hashed features. The $C$ parameter of SVM was tuned using cross-validation, for every individual run, to ensure the best possible performance on every combination of the number of features and the hashing scheme. This ensures fairness of the comparisons.

Figure~\ref{fig:class} compares the mean average precision of classification tasks using Densified WTA codes, WTA codes and basic sparse BoW on three data sets. The baseline, mean average precision for the three BoW with different bins are shown by dashed straight lines. The mean average precision for WTA feature vectors are shown by dot-dashed curves and for Densified WTA feature vectors are shown by dot-dashed curves. We could see that as stated in~\cite{yagnik2011power}, precision increases when original BoW bin number increases or the number of codes increases with WTA beating BoW in each case. These observations are in line with the original WTA paper. We followed the experiment pipeline from the WTA paper, while generating BoW using standard package~\cite{vedaldi2010vlfeat}. It is not surprising to see exactly same trends in classification results with difference in relative values.

The Densified WTA consistently outperforms Vanilla WTA significantly on all the three datasets, irrespective of the choice of BoW or number of hashes. Moreover, the performance gap increases with the number of BoW. The increase in BoW increases sparsity of the dataset and hence this trend clearly validates our hypothesis and the theory in this paper. The gains over WTA are significant and our results clearly push the boundary of classification performance with hashing based kernels significantly outperforming BoW. Note that increasing BoW from 5000 to 10000 leads to no gains in accuracy. But with hashing, especially Densified WTA, the gains keep climbing.

\subsection{Image Retrieval }

We now compare the performance of our Densified WTA codes with Vanilla WTA by replicating the retrieval experiments and studying the standard precision recall
curves. This is our main task of performance comparison because like we mentioned in Section~\ref{section1}, WTA is quite appealing for information retreival.

For each query image, the nearest-neighbors of each test data were ranked among training data based on Hamming distance of the hash codes. Since we had labeled datasets, all the images with the same label as the query were treated as the gold standard neighbors. Note, as mentioned in Section~\ref{others} WTA and Densified WTA leads to two different similarity measure (or kernel). Therefore, this experiment is comparing which among these two kernels agrees with the ground truth labels.

We again choose K=4, which was picked using the same method described in~\cite{yagnik2011power} and best for WTA. The precision and recall curves for the rankings based on different hash codes are shown in Figure~\ref{fig:ret}. We show plots for 64, 256 and 512 hash codes on
1000, 5000 and 10000 BoW representations (9 curves for each dataset per hashing scheme). To average out the randomness of both Densified and original WTA hashing, every curve on the graphs is averaged from 10 runs.

We can see that Densified WTA hashes lead to notably better precision-recall compared to Vanilla WTA on all combinations irrespective of the choices of the dataset. As with classification, an increase in BoW leads to larger gap due to increasing in sparsity. This again validates our claims in the paper.

It is very exciting to see that a small but principled modification to WTA Hashing can lead to drastic benefits.

\section{Conclusions}
In this work, we revisited the problem of WTA Hashing for very sparse datasets which are ubiquitous in large scale applications. We found a particular issue with WTA hashing in this regime which makes them uninformative with an increase in sparsity. We provide a principled solution to this problem using the novel idea of ``Densification".  Our solutions leverage the theoretical benefits of rank correlation methods and at the same time successfully resolves the concern of uninformative hash values produced by WTA Hashing for data with high sparsity. Evaluation results shown confirm the superior performance of Densified WTA Hashing on both image classification and retrieval task.

\bibliographystyle{abbrv}
\bibliography{main_bib}

\begin{thebibliography}{10}

\bibitem{msrc}
Microsoft research cambridge object recognition image database.
\newblock Microsoft Research, 2004.

\bibitem{canini2012sibyl}
K.~Canini, T.~Chandra, E.~Ie, J.~McFadden, K.~Goldman, M.~Gunter, J.~Harmsen,
  K.~LeFevre, D.~Lepikhin, T.~Llinares, et~al.
\newblock Sibyl: A system for large scale supervised machine learning.
\newblock {\em Technical Talk}, 2012.

\bibitem{chapelle1999support}
O.~Chapelle, P.~Haffner, and V.~N. Vapnik.
\newblock Support vector machines for histogram-based image classification.
\newblock {\em Neural Networks, IEEE Transactions on}, 10(5):1055--1064, 1999.

\bibitem{dean2013fast}
T.~Dean, M.~Ruzon, M.~Segal, J.~Shlens, S.~Vijayanarasimhan, and J.~Yagnik.
\newblock Fast, accurate detection of 100,000 object classes on a single
  machine.
\newblock In {\em Proceedings of the IEEE Conference on Computer Vision and
  Pattern Recognition}, pages 1814--1821, 2013.

\bibitem{everingham2010pascal}
M.~Everingham and J.~Winn.
\newblock The pascal visual object classes challenge 2010 (voc2010) development
  kit, 2010.

\bibitem{indyk1998approximate}
P.~Indyk and R.~Motwani.
\newblock Approximate nearest neighbors: towards removing the curse of
  dimensionality.
\newblock In {\em Proceedings of the thirtieth annual ACM symposium on Theory
  of computing}, pages 604--613. ACM, 1998.

\bibitem{jiang2007floating}
A.~Jiang, V.~Bohossian, and J.~Bruck.
\newblock Floating codes for joint information storage in write asymmetric
  memories.
\newblock In {\em Information Theory, 2007. ISIT 2007. IEEE International
  Symposium on}, pages 1166--1170. IEEE, 2007.

\bibitem{kokkinos2012dense}
I.~Kokkinos, M.~Bronstein, and A.~Yuille.
\newblock Dense scale invariant descriptors for images and surfaces.
\newblock 2012.

\bibitem{lee2006efficient}
H.~Lee, A.~Battle, R.~Raina, and A.~Y. Ng.
\newblock Efficient sparse coding algorithms.
\newblock In {\em Advances in neural information processing systems}, pages
  801--808, 2006.

\bibitem{Article:Li_Konig_CACM11}
P.~Li and A.~C. K\"onig.
\newblock Theory and applications b-bit minwise hashing.
\newblock {\em Commun. ACM}, 2011.

\bibitem{li2011hashing}
P.~Li, A.~Shrivastava, J.~L. Moore, and A.~C. K{\"o}nig.
\newblock Hashing algorithms for large-scale learning.
\newblock In {\em Advances in neural information processing systems}, pages
  2672--2680, 2011.

\bibitem{parikh2011relative}
D.~Parikh and K.~Grauman.
\newblock Relative attributes.
\newblock In {\em Computer Vision (ICCV), 2011 IEEE International Conference
  on}, pages 503--510. IEEE, 2011.

\bibitem{Proc:Rahimi_NIPS07}
A.~Rahimi and B.~Recht.
\newblock Random features for large-scale kernel machines.
\newblock In {\em Advances in neural information processing systems}, pages
  1177--1184, 2007.

\bibitem{russell2008labelme}
B.~C. Russell, A.~Torralba, K.~P. Murphy, and W.~T. Freeman.
\newblock Labelme: a database and web-based tool for image annotation.
\newblock {\em International journal of computer vision}, 77(1-3):157--173,
  2008.

\bibitem{shrivastava2015probabilistic}
A.~Shrivastava.
\newblock {\em Probabilistic Hashing Techniques For Big Data}.
\newblock PhD thesis, Cornell University, 2015.

\bibitem{shrivastava2014densifying}
A.~Shrivastava and P.~Li.
\newblock Densifying one permutation hashing via rotation for fast near
  neighbor search.
\newblock In {\em Proceedings of the 31st International Conference on Machine
  Learning (ICML-14)}, pages 557--565, 2014.

\bibitem{Proc:Shrivastava_UAI14}
A.~Shrivastava and P.~Li.
\newblock Improved densification of one permutation hashing.
\newblock In {\em UAI}, Quebec, CA, 2014.

\bibitem{vedaldi2010vlfeat}
A.~Vedaldi and B.~Fulkerson.
\newblock Vlfeat: An open and portable library of computer vision algorithms.
\newblock In {\em Proceedings of the 18th ACM international conference on
  Multimedia}, pages 1469--1472. ACM, 2010.

\bibitem{Proc:Weber_VLDB98}
R.~Weber, H.-J. Schek, and S.~Blott.
\newblock A quantitative analysis and performance study for similarity-search
  methods in high-dimensional spaces.
\newblock In {\em Proceedings of the 24rd International Conference on Very
  Large Data Bases}, VLDB '98, pages 194--205, San Francisco, CA, USA, 1998.
  Morgan Kaufmann Publishers Inc.

\bibitem{yagnik2011power}
J.~Yagnik, D.~Strelow, D.~A. Ross, and R.-s. Lin.
\newblock The power of comparative reasoning.
\newblock In {\em Computer Vision (ICCV), 2011 IEEE International Conference
  on}, pages 2431--2438. IEEE, 2011.

\bibitem{yang2009linear}
J.~Yang, K.~Yu, Y.~Gong, and T.~Huang.
\newblock Linear spatial pyramid matching using sparse coding for image
  classification.
\newblock In {\em Computer Vision and Pattern Recognition, 2009. CVPR 2009.
  IEEE Conference on}, pages 1794--1801. IEEE, 2009.

\bibitem{Ziegler2012NIPS}
A.~Ziegler, E.~Christiansen, D.~Kriegman, and S.~Belongie.
\newblock Locally uniform comparison image descriptor.
\newblock In {\em Neural Information Processing Systems (NIPS)}, pages 1--9,
  Lake Tahoe, NV, 2012.

\end{thebibliography}
\end{document}